\pdfoutput=1

\documentclass[11pt]{article}

\usepackage[]{EMNLP2022}

\usepackage{times}
\usepackage{latexsym}

\usepackage[T1]{fontenc}

\usepackage[utf8]{inputenc}

\usepackage{microtype}

\usepackage{inconsolata}
\usepackage{amsmath,graphicx}
\usepackage{subfig}
\usepackage{booktabs}
\newcommand{\prefix}{\textcolor{black}{\textsc{PF}}}
\newcommand{\adapter}{\textcolor{black}{\textsc{AP}}}
\newcommand{\finetune}{\textcolor{black}{\textsc{FT}}}
\newcommand{\promptuning}{\textcolor{black}{\textsc{PT}}}
\newcommand{\PERM}{\textcolor{black}{\textsc{PERM}s}}

\title{Evaluating Parameter Efficient Learning for Generation}

  \author{Peng Xu$^{\mathsection}$, Mostofa Patwary$^{\mathsection}$, Shrimai Prabhumoye$^{\mathsection}$, Virginia Adams$^{\mathsection}$, Ryan J. Prenger$^{\mathsection}$,   \\
  \textbf{Wei Ping$^{\mathsection}$, Nayeon Lee$^{\ddagger}$, Mohammad Shoeybi$^{\mathsection}$, Bryan Catanzaro$^{\mathsection}$} \\
  $^{\ddagger}$The Hong Kong University of Science and Technology,
  $^{\mathsection}$NVIDIA \\
  \texttt{pengx@nvidia.com} \\}

\begin{document}
\maketitle
\begin{abstract}
Parameter efficient learning methods ({\PERM}) have recently gained significant attention as they provide an efficient way for pre-trained language models (PLMs) to adapt to a downstream task. 
However, these conclusions are mostly drawn from in-domain evaluations over the full training set. In this paper, we present comparisons between {\PERM} and finetuning from three new perspectives: (1) the effect of sample and model size to in-domain evaluations, (2) generalization to unseen domains and new datasets, and (3) the faithfulness of generations.
Our results show that for in-domain settings (a) there is a cross point of sample size for which {\PERM} will perform better than finetuning when training with fewer samples, and (b) larger PLMs have larger cross points. 
For cross-domain and cross-dataset cases, we show that (a) Adapter \cite{houlsby2019parameter} performs the best amongst all the {\PERM} studied here, and (b) it outperforms finetuning if the task dataset is below a certain size.
We also compare the faithfulness of generations and show that {\PERM} can achieve better faithfulness score than finetuning, especially for small training set, by as much as 6\%. Finally, we apply Adapter to MT-NLG 530b \cite{smith2022using} and  achieve new state-of-the-art results on Xsum \cite{narayan2018don} for all ROUGE scores (ROUGE-1 49.17, ROUGE-2 27.20, ROUGE-L 40.98).
\end{abstract}

\section{Introduction}



Parameter efficient learning methods ({\PERM}) serve as potential alternatives to finetuning for adapting and deploying language models in real world scenarios \cite{ding2022delta}.  They allow users to finetune only a small number of parameters while freezing the rest of the shared parameters of pre-trained language models (PLMs). This is especially important for large language models (e.g. GPT-3 \cite{brown2020language} and MT-NLG \cite{smith2022using}) 
as finetuning the entire model will be very expensive or infeasible due to their model size.

Prefix tuning \cite{li2021prefix}, which is one of the {\PERM}, draws inspiration from prompting and introduces a small set of continuous vectors as virtual prompts to allow subsequent tokens to attend to, which obtains comparable performance to finetuning in the full data setting.  Prompt tuning \cite{lester2021power} shows the power of scaling PLMs and  that tuning only a few extra embeddings is sufficient to achieve similar performance to finetuning the entire 11b T5-XXL \cite{raffel2020exploring} model. 
P-tuning v2 \cite{liu2022p} further demonstrates that small PLMs can also achieve comparable results to finetuning with Prefix tuning. Different from adding new parameters through prompts, Adapter \cite{houlsby2019parameter} injects trainable parameters through low-rank structure in a skip-connection way. Other {\PERM} includes LoRA \cite{hu2021lora}, Mix-And-Match adapter \cite{he2021towards}, Compactor \cite{karimi2021compacter}, BitFit \cite{zaken2022bitfit}, diff-pruning \cite{guo2021parameter} and etc. 

Most conclusions about {\PERM} so far are drawn from their in-domain evaluations over full training samples. To the best of our knowledge, it is not yet investigated (1) how these conclusions apply to different training sizes and model sizes, and (2) how {\PERM} generalize to unseen domains and new datasets, which are both important aspects for deploying {\PERM} in real-world applications. 

In addition, faithfulness in natural language generation has become an important topic as it is vital to real-world applications. Various efforts are made to systematically measure and mitigate factual errors in many generation tasks, including summarization~\cite{huang2021factual} and dialogue generations~\cite{rashkin2021increasing,shuster2021retrieval,dziri2021neural, wu2021controllable}. However, existing work on faithfulness only focuses on faithfulness of finetuning, and the impact of {\PERM} on the faithfulness of generation is not yet explored.

In this paper, we provide an in-depth study of {\PERM} for generation tasks through three important aspects when deploying {\PERM} in practical applications: (1) in-domain evaluation by scaling both training dataset size and model size of PLMs, (2) cross-domain and cross-dataset generalization, and (3) faithfulness assessment. Two generation tasks are used for evaluation: summarization and dialogue generation. We study four representative methods: P-tuning, Prompt tuning, Prefix tuning, and Adapter, but mainly focus on Prefix tuning and Adapter as our preliminary results show that they are better than the others. 
Our contributions are summarized as follows: (1) To the best of our knowledge, we present the first comparisons of faithfulness for {\PERM}. Our experimental results show that {\PERM}, especially prefix tuning can achieve better faithfulness than finetuning by up to 6\%.  (2) For in-domain settings, there is always a cross point of sample size for which {\PERM} will be better than finetuning when training on fewer samples. Larger PLMs have larger cross points. Users need to choose which method to use based on their own training sample size and model size. (3) Compared to finetuning, not all {\PERM} can easily achieve better cross-domain and cross-dataset scores than finetuning even with 8.3b PLM. Our results show that Adapter is a better method than Prefix tuning on 13 out of 15 comparison settings. (4) New state-of-the-art results on Xsum \cite{narayan2018don} are obtained by applying Adapter to MT-NLG 530b model. 

\section{Methodology}




We compare the following four {\PERM} to \textbf{finetuning} (\textbf{\finetune}) using GPT-style models from Megatron-LM \cite{shoeybi2019megatron}. 
\textbf{(1) Adapter (\adapter)} adds an extra layer with a bottleneck structure by first projecting input $h$ to a low dimension using trainable weights $W_{down}$ and then projecting up to the original dimension using trainable weights $W_{up}$. It is incorporated into backbone model in a skip-connection way. 
\begin{equation*}
    \textit{Adapter}(h) = h + g(h W_{down}) W_{up},
\end{equation*}
where $g$ is the activation function.
In our case, we insert \textit{Adapter} layer both after the multi-head attention (MHA) and feedforward layer (FFD) of Transformer \cite{vaswani2017attention}.
\textbf{(2) Prefix Tuning (\prefix)} adds trainable prefix tokens at the beginning of each transformer block. We follow the implementation of \citet{li2021prefix} to replace the keys $K$, values $V$ of MHA with the concatenation of the trainable prefix weights $W_{K}$, $W_{V}$ and the $K, V$. 
\begin{align*}
    K & \leftarrow \textit{concat}([W_{K} ; K ]) \\
    V & \leftarrow \textit{concat}([W_{V} ; V ])
\end{align*}
We also add reparameterization trick suggested by \citet{li2021prefix}. 
\textbf{(3) Prompt Tuning (\promptuning)} adds extra parameters to the embedding layer and uses these trainable embeddings to prompt the input. 
\textbf{(4) P-tuning \cite{liu2021gpt}} adds a prompt encoder to encode pseudo prompts and  the encoded representation is used to prompt the input.


\section{Experimental Setup}

\subsection{Datasets}
\paragraph{Summarization} We use Xsum \cite{narayan2018don}, a widely used summarization dataset, to train and evaluate different methods. It consists of 204,017/11,327/11,333 pairs for the training/validation/test. 
As Xsum does not divide the dataset based on topics,  we follow \citet{li2021prefix} to split the Xsum dataset into news articles for training and sports articles for testing. This cross-domain version has 149,115/8,263/2,823 pairs for training/validation/test.  For the cross-dataset evaluation, we choose the test set from CNN/Daily Mail \cite{nallapati2016abstractive}. It contains 11,490 samples.

\paragraph{Dialogue} We use Wizard of Wazards (WoW) \cite{dinan2018wizard} dataset for our dialogue generation task. The modeling of the wizard response is usually composed of two steps: knowledge retrieval and response generation. To simplify the problem, following \citet{rashkin2021increasing}, we ignore the knowledge retrieval step and take the golden knowledge for the response generation. The response of the wizard is then used to train the model. For the cross-dataset evaluation, we use the CMU\_DoG \cite{zhou2018dataset} dataset. We test our model over all test set dialogue turns except the starting one.





\subsection{Metrics}

\paragraph{Quality Metrics} 
We use ROUGE-1 (R-1), ROUGE-2 (R-2), ROUGE-L (R-L) \cite{lin2004rouge} scores to evaluate the generations for summarization task as it is well adopted in all summarization tasks. For the dialogue generation task, \citet{dinan2018wizard} reports both PPL and unigram F1 scores and better PPL in general gives better F1 scores. \citet{adiwardana2020towards} also shows high correlation between PPL and the quality of dialogue based on human evaluations. We therefore choose to report PPL as an indicator of the quality of generated dialogues. In the Results section, if we say ``A is better than B”, we mean A has a higher ROUGE score for summarization task or/and a lower PPL score for dialogue tasks. 

\paragraph{Faithfulness Metrics}
Following \citet{rashkin2021increasing}, we use a state-of-the-art natural language interference (NLI) model
(Roberta trained on MNLI \cite{liu2019roberta}) to predict whether a response can be entailed by the given evidence. We evaluate the faithfulness of generated response against the concatenation of dialogue history and the golden knowledge.  Entailment score is reported as the ratio of the samples being predicted as entailment from the NLI model. We use factCC \cite{kryscinski2020evaluating} to evaluate the faithfulness for the Xsum  as it has the highest Spearman correlation with human evaluations \cite{pagnoni2021understanding}. 

\section{Results}

\begin{table}[t]
\centering
\scalebox{0.89}{
\begin{tabular}{cccccc}
\hline
Method  &Parameter		& R-1 & R-2  & R-L\\ \hline 
P-tuning  & 72k  & 33.3 & 11.2 & 26.0 \\
\promptuning & 154k   & 32.7 & 10.8 & 25.5 \\
\prefix & 5m & 35.3 & 13.5 & 27.9  \\
\adapter  & 5m  & \bf 37.7  & \bf 15.3 & \bf 30.1 \\ \hline 
\finetune	 & 357m & \bf 41.6 &\bf  19.2  &\bf  33.8  \\ \hline 
\end{tabular}}
\caption{Xsum results by comparing different methods over 357m GPT model using full dataset. Parameter here counts extra task  parameters needed during inference. {\finetune} is much better than {\PERM} for all ROUGE metrics with p-value lower than 0.001 through a t-test. {\adapter} and {\prefix} is also better than P-tuning and PT with p-value lower than 0.001 through a t-test.}
\label{tab:xsum-all-method-comp}
\end{table}

\begin{figure}[t]
\begin{center}
\subfloat[R-L comparison over Xsum test set]{%
  \includegraphics[width=\columnwidth]{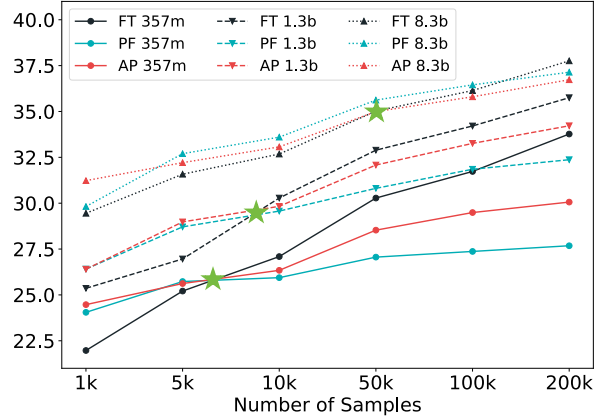}%
  \label{fig:xsum-in-domain}
}

\subfloat[PPL comparison over WoW seen test set]{%
  \includegraphics[width=\columnwidth]{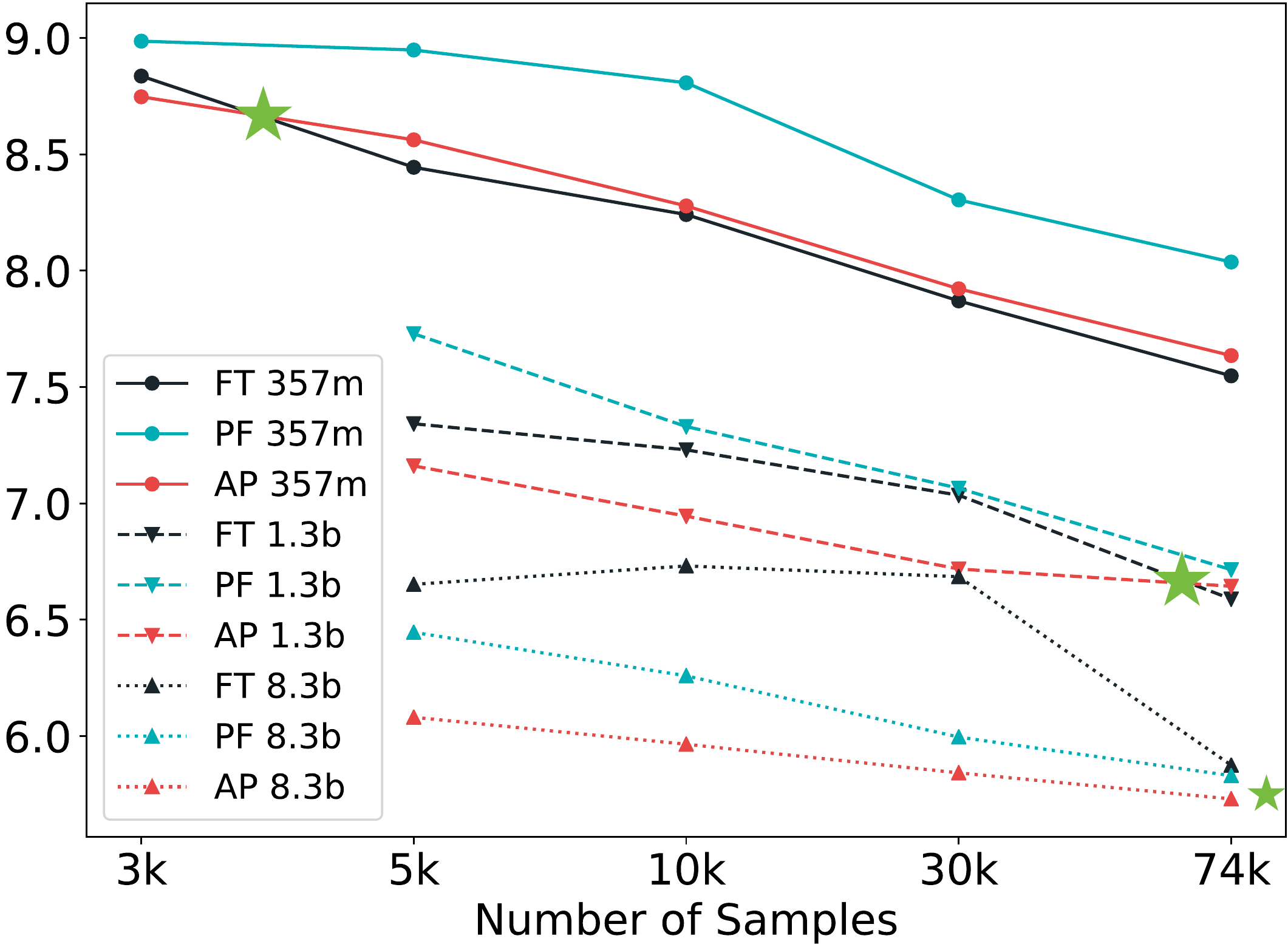}%
  \label{fig:wow-in-domain}
}

\end{center}
\caption{ There is a cross point of sample size where {\PERM}, e.g. {\adapter} will be better than {\finetune} when training fewer samples. Larger PLMs have larger cross points.}
\label{fig:in-domain}  
\end{figure}

\subsection{In-domain Results}
In-domain evaluations are presented in Table \ref{tab:xsum-all-method-comp}. Although Adapter({\adapter}) and Prefix Tuning ({\prefix}) are better than prompt tuning and p-tuning, they are still much worse than {\finetune} (3.7 lower for R-L). 
To better understand when {\PERM} is better than {\finetune}, we scale both the training sample size and model size for summarization and dialogue generation task. As  {\prefix} and {\adapter} are much better than other {\PERM}, we focus on those two methods. 

The results on Xsum and WoW are shown in Figure \ref{fig:xsum-in-domain} and Figure \ref{fig:wow-in-domain}. Comparing {\adapter} with {\prefix}, we find that {\adapter} is better than {\prefix} on 26 out of 31 comparisons. It is also aligned with the conclusion in \citet{ding2022delta}. This can be attributed to the structural bias of Adapter. The skip-connection structure allows Adapter to add a small deviation to the activation, which makes the optimization of the PLM checkpoint smooth. On the contrary, {\prefix} introduces deviations to the keys and values of the self-attention module and therefore greatly varies the activation of each layer. As a result, it takes more efforts for {\prefix} to converge. Another phenomenon we observed in Figure \ref{fig:in-domain} is that if we train a 8.3b model with enough training samples (74k for WoW or 200k for Xsum), the performance gap between {\prefix}, {\adapter} and {\finetune} is quite marginal. This suggests us to use {\PERM} instead of {\finetune} to save the cost of deploying 8.3b PLMs. 

Comparing {\finetune} with {\adapter}, we find there is always a cross point of sample size where {\finetune} is better than {\adapter}. This shows that if we have large number of samples in training set, {\finetune} will work better. But if the number of samples for the task 
are small, {\adapter} will be better. Also, this cross point will be larger if we use larger PLMs. For example, the cross point for 1.3b model over Xsum is less than 10k samples, whereas for the 8.3b model, it is 50k samples . This phenomenon can be attributed to that {\finetune} can easily overfit when you have large models or few training samples. It motivates us to use {\adapter} when you have small dataset or large model to achieve better in-domain performances. 

Interestingly, tuning {\adapter} with a 8.3b model of only 32m extra parameters over 5k samples achieves much better results than finetuning 357m model over 100k samples. This means more than 90\% task specific parameters can be saved for deployment and more than 97\% tasks specific samples can be reduced for training by sharing the larger frozen PLMs.

\begin{table}[t]
\centering
\scalebox{0.9}{
\begin{tabular}{cccccc}
\hline
Method   & Parameter		& R-1 & R-2  & R-L\\ \hline 
BRIO & 568m  &  49.07 & 25.59 & 40.40 \\
T5 & 11b   &  48.83 & 25.96 &  40.70 \\
MT-NLG & 103m & 49.17 & 27.20 & 40.98  \\\hline
\end{tabular}}
\caption{Xsum results by comparing MT-NLG {\adapter} to other state-of-the-art models: (1) BRIO \cite{liu2022brio} (2) T5 \cite{rothe-etal-2021-thorough}. MT-NLG achieves new state-of-the-art results.}
\label{tab:xsum-sota-comp}
\end{table}

\paragraph{Scaling up to 530b model} Since {\adapter} gets better performances than other methods, we apply {\adapter} to one of the largest GPT model, MT-NLG \cite{smith2022using}. Table \ref{tab:xsum-sota-comp} shows that by adding 103m parameters to a frozen 530b model, we achieves a new state-of-the-art result on Xsum for all ROUGE scores (R-1 49.17, R-2 27.20, R-L 40.98). It outperforms both strong encoder-decoder models (e.g. T5 \cite{rothe-etal-2021-thorough}), as well as specialized models (e.g. BRIO \cite{liu2022brio}). This result shows that decoder-only  model can still beat encoder-decoder  model, but it needs a much larger model size.

\begin{table}[t]
\centering
\scalebox{0.95}{
\begin{tabular}{ccccc}
\hline
model size	& parameters	&	\prefix	& \adapter \\ \hline
1.3b	& 5m &  32.76 & 33.73  \\
1.3b	& 10m & 32.37 & 34.22 \\ \hline
8.3b    & 5m & 37.21 & 37.71 \\ 
8.3b	& 33m  & 37.14  & 36.73\\ \hline
\end{tabular}}
\caption{R-L score for {\prefix} and {\adapter}. More parameters do not always give better results.} 
\label{tab:xsum-param-not-help}
\end{table}

\begin{table}[t]
\centering
\scalebox{0.95}{
\begin{tabular}{ccccc}
\hline
model size	& dataset	&	\prefix	& \adapter & 
\finetune \\ \hline
357m	& WoW &  8.35 & 8.01 & \textbf{7.94} \\
1.3b	& WoW & 6.99 & 6.91 & \textbf{6.89} \\ 
8.3b & WoW & 6.06	& \textbf{5.95}	& 6.11 \\ \hline 
357m	& Xsum & 23.86 & \textbf{24.55}	& 23.07	 \\
1.3b	& Xsum & \textbf{28.85} & 28.60 & 27.74 \\ \hline
\end{tabular}}
\caption{Cross-domain evaluation with R-L and PPL over Xsum and WoW trained with full dataset samples.}
\label{tab:cross-domain}
\end{table}

\begin{table}[t]
\centering
\scalebox{0.95}{
\begin{tabular}{ccccc}
\hline
model size	& samples	&	\prefix	& \adapter & \finetune \\ \hline
357m	& 5k  & 18.70 & \textbf{19.91} & 17.11\\
1.3b	& 5k &  19.19 & \textbf{19.22} & 17.97 \\ \hline 
357m	&200k & \textbf{17.84} & 16.54 & 14.00 \\
1.3b	& 200k & 15.43 & \textbf{15.81} & 14.20 \\ \hline
\end{tabular}}
\caption{Cross-dataset generalization evaluation over CNN/Daily Mail using R-L. {\PERM} outperforms {\finetune}. }
\label{tab:cross-dataset-cnndaily}
\end{table}

\begin{table}[t]
\centering
\scalebox{0.95}{
\begin{tabular}{ccccc}
\hline
model size	& samples	&	\prefix	& \adapter & \finetune \\ \hline
357m	& 5k	& 29.7	& \textbf{26.7}	& 27.1 \\
1.3b	& 5k	& 26.3	& \textbf{19.7}	& 21.0 \\
8.3b	& 5k	& 16.9	& \textbf{15.3} & 17.5 \\ \hline 
357m	& 74k	& 30.0	& 27.6	& \textbf{26.2} \\
1.3b	& 74k	& 24.0	& \textbf{20.1}	& 21.0 \\
8.3b	& 74k	& 17.8	& \textbf{16.3}	& 16.5 \\ \hline
\end{tabular}}
\caption{Cross-dataset generalization evaluation over CMU\_DoG using PPL.}
\label{tab:cross-dataset-cmudog}
\end{table}

We also study the effects of varying parameter sizes for {\PERM} in Table \ref{tab:xsum-param-not-help}. We found that the score of AP increases for 1.3b model, which suggests the model is under-fitting.  \citet{he2021towards} observed a similar trend with a similar sized model (700M). On the other hand, the score of {\prefix} drops for 1.3b model, which suggests it is overfitting. This difference can be attributed to the way we count the parameters in Table \ref{tab:xsum-param-not-help}. Note that the number of parameters for {\prefix} is counted as extra \textit{inference} parameters following \citet{li2021prefix}, which is different from trainable parameters. For example, {\prefix} 1.3b model with 10m extra inference parameters actually contains 80m extra trainable parameters, which is much higher than the 10m shown in the table. Such a large number of trainable parameters will easily make the model overfit for {\prefix} and thus leads to the performance drop with more parameters. For the 8.3b model, the scores of both {\adapter} and {\prefix} drops with more parameters as both of them are overfitting and AP has a more serious overfitting issue there. 
Table \ref{tab:xsum-param-not-help} suggests that (1) more parameters do not always help {\prefix} or {\adapter}, (2) task specific parameters can be further reduced by sacrificing little scores.

\subsection{Cross-domain and Cross-dataset Generalization}

Table \ref{tab:cross-domain} shows cross-domain results over Xsum and WoW and cross-dataset results can be found in Table \ref{tab:cross-dataset-cnndaily} and Table \ref{tab:cross-dataset-cmudog}.  We find that {\adapter} achieves in general better generalization than  {\prefix} in cross-domain and cross-dataset setting by 13 out of 15 comparisons.

For Xsum, both {\adapter} and {\prefix} are universally better than {\finetune} for cross-domain and cross-dataset setting across different training sample sizes. For WoW, we find the conclusion is a bit different. {\prefix} is worse than {\finetune} with 7 out of 9 comparison settings in Table \ref{tab:cross-domain} and Table \ref{tab:cross-dataset-cmudog} while {\adapter} has only 3 out of 9. We conjecture this is due to the bias of different tasks as we can see the cross points for Xsum and WoW are also quite different. WoW reflects a more general finding  that {\adapter} can achieve better cross-domain and cross-dataset scores than {\finetune} when PLMs are large enough (e.g. 1.3b) or training samples are small enough (e.g. 5k). In these two cases, {\adapter} only tunes a relatively small amount of parameters comparing to {\finetune} and therefore is less likely to overfit. It is also aligned with conclusions under in-domain scenario.



\subsection{Faithfulness}
Table \ref{tab:faith-dialogue} and Table \ref{tab:faith-xsum} shows the faithfulness evaluation over WoW and Xsum dataset using entailment score and factCC score. The faithfulness score for Xsum is quite low as the dataset contains many unfaithful training samples \cite{pagnoni2021understanding}. Both of the tables show that {\prefix} achieves the best faithfulness score across all model sizes and sample size. However, when increasing the PLM size from 357m to 8.3b, or training samples from 5k to 74k, we see a constant drop of entailment score or factCC score. This can be attributed to (1) both WoW and Xsum have many responses or summaries that contains information external to the evidence \cite{rashkin2021increasing, pagnoni2021understanding} and (2) larger language models memorize more world knowledge itself \cite{brown2020language}. Therefore, our models will become more unfaithful when they learn from those unfaithful examples or use its embedded knowledge. In such case, {\prefix} provides an option for users to sacrifices a little PPL to earn more faithfulness. 
How to further improve the faithfulness of {\PERM} is still an open research problem and we leave it for future work.

\begin{table}[t]
\centering
\scalebox{0.95}{
\begin{tabular}{ccccc}
\hline
model size	& samples	&	\prefix	& \adapter & \finetune \\ \hline
357m	& 5k	& \textbf{0.815}	& 0.800	 & 0.751 \\
1.3b	& 5k	& \textbf{0.768}	& 0.749	 & 0.713 \\
8.3b	& 5k	& \textbf{0.752}	& 0.733	 & 0.700  \\  \hline 
357m	& 74k	& 	\textbf{0.788}   & 0.767	& 0.762 \\
1.3b	& 74k	& 	\textbf{0.760}	& 0.744	& 0.750 \\
8.3b	& 74k	& 	\textbf{0.721}	& 0.705	& 0.720 \\ \hline

\end{tabular}}

\caption{Entailment score for WoW test seen dataset. {\prefix} achieves the best score.}
\label{tab:faith-dialogue}
\end{table}

\begin{table}[t]
\centering
\scalebox{0.95}{
\begin{tabular}{ccccc}
\hline
model size 	& samples	&	\prefix	& \adapter & \finetune \\ \hline
357m	& 200k   & \textbf{0.252}	& 0.239	& 0.232\\
1.3b	& 200k	 & \textbf{0.243}	& 0.241	& 0.227\\
8.3b	& 200k	 & \textbf{0.231}	& 0.219	& 0.227 \\ \hline
\end{tabular}}
\caption{FactCC scores for Xsum. {\prefix} achieves the best score.}
\label{tab:faith-xsum}
\end{table}

\section{Conclusion}
In this paper, we extensively compare {\PERM} with finetuning over three main areas: (1) in-domain evaluation  by scaling both the sample size and model size  (2) cross-domain and cross-dataset generalization (3) faithfulness of generations. For in-domain settings, we find (a) there is a cross point of sample size for which parameter efficient learning will be better than finetuning when training with fewer samples and (b) larger PLMs have larger cross points. This suggests users to choose the method based on their own sample size and model size. Simply apply Adapter to MT-NLG, we achieve new state-of-the-art results on Xsum for all ROUGE scores (ROUGE-1 49.17, ROUGE-2 27.20, ROUGE-L 40.98). Compared to finetuning, not all {\PERM} can easily achieve better cross-domain and cross-dataset scores than finetuning even with large PLM (e.g. 8.3b). Adapter is a better choice than other {\PERM} in such cases. Lastly, we provide the first comparison of {\PERM} over faithfulness of generations and show that Prefix tuning is the best method for faithfulness. We believe our findings will help users better choose and deploy {\PERM}.

\section{Limitations}
Our paper have the following limitations. Firstly, we are only able to qualitatively show the cross point when {\finetune} is better than {\adapter}. We do not derive a quantitative estimation of the cross point given the model size and the task name. Therefore, the cross point of our paper can be served as a reference only for summarization and dialogue generation when choosing between these methods. (2) Even though we show {\prefix} achieves better faithfulness than other methods. We found that when the model is large enough, (e.g. 8.3b) and the dataset is large too (e.g. 74k), {\prefix} achieves quite close scores to {\finetune} (0.721 vs 0.720). Therefore, it remains a question how to achieve better faithfulness under such setting. 

\bibliography{anthology,custom}
\bibliographystyle{acl_natbib}

\clearpage
\appendix

\section{Example Appendix}
\label{sec:appendix}

\subsection{Training Details}
We use the pre-trained GPT checkpoint trained by Megatron-LM \cite{shoeybi2019megatron}. 
When number of samples is less than 5000, we set batch size as 8 and otherwise we set it as 64. We set learning rate as 1e-4 for Xsum dataset when running Adapter (\adapter) tuning or Prefix tuning (\prefix). We set it as 3e-4 for WoW dataset. For finetuning over Xsum dataset, we use the learning rate of 3e-5 for 357m and 1e-5 for 1.3b and 8.3b model. When finetuning WoW dataset, we set it for 8e-6 for all model sizes. Xsum dataset is trained for 10 epochs and WoW dataset is trained for 5 epochs. We used ROUGE-L score at the validation set to select the best model for Xsum and used PPL at the seen validation set for WoW. To {\adapter} tune 530b MT-NLG model, we set the learning rate as 3e-5 and the batch size as 32.

The prefix length is fixed as 100 and hidden dimension as 800 for {\prefix} experiments \cite{li2021prefix}. For {\adapter}, the hidden size was set to 50 to achieve a similar extra number of parameters for inference.  We summarize the extra task specific parameters introduced by {\prefix} and {\adapter} in Table \ref{tab:extra-params}. Note that we don't intend to do extensive hyperparameter search for all the combinations for model size, sample size and tuning methods. We instead would like to draw conclusions that can generalize across model size, sample size and tasks. We used NVIDIA V100 and A100 GPUs to run all experiments.

\begin{table}[h]
\centering
\scalebox{0.89}{
\begin{tabular}{ccc}
\toprule
Model	& Method  & Parameter \\	 \midrule
357m	& 	P-tuning  & 72k   \\
357m	& 	\promptuning & 154k   \\
357m	& 	\prefix & 5m   \\
357m	& 	\adapter  & 5m   \\ 
1.3b	& 	\prefix & 10m   \\
1.3b	& 	\adapter  & 10m   \\ 
8.3b	& 	\prefix & 33m   \\
8.3b	& 	\adapter  & 33m   \\ \bottomrule
\end{tabular}}
\caption{Extra parameters for different methods}
\label{tab:extra-params}
\end{table}

For summarization, we simply give the article as the input and the summary as the output. For dialogue, We formulate the input with the following template: 
$``\{\text{TOPIC}\} \backslash t \{\text{dialogue\_history}\} \backslash t \text{Knowledge:} \\ \{\text{knowledge}\} \backslash t  \backslash t  \text{B:}"$. For dialogue history, we add $A$: and $B$: in front of each utterance to distinguish different speakers. $\backslash t$ is used to separate different dialogue turns. We use beam search for the decoding step and we set beam size as 5 for all settings.


\subsection{More Related Work}

More work about faithfulness include summarization~\cite{cao2018faithful,dong2020multi,cao2021cliff,huang2021factual}, dialogue generations~\cite{rashkin2021increasing,shuster2021retrieval,dziri2021neural, wu2021controllable}, data-to-text~\citep{wiseman2017challenges,nie2019simple,liu2021towards,rebuffel2022controlling}, and translation~\cite{lee2018hallucinations,wang2020exposure}. However, still relatively little is known about faithfulness/hallucination problem. \citeauthor{pagnoni2021understanding} conduct a good analysis of error types. 

Other parameter efficient learning methods includes  PPT \cite{gu2022ppt}, masking \cite{zhao2020masking} with application in  multitask learning \cite{stickland2019bert, mahabadi2021parameter}, transfer learning \cite{pfeiffer2021adapterfusion, su2022transferability, vu2022spot}, improving robustness \cite{han2021robust}, low resources settings \cite{he2021effectiveness}

\subsection{Additional Results}

We present detailed ROUGE scores for Xsum in the following tables.

\begin{table*}[t]
\centering
\small{
\begin{tabular}{@{}c c c c c @{\hskip 0.35in} c c c @{\hskip 0.35in} c c c@{}}
\textbf{Model} & \textbf{Samples} & \multicolumn{3}{c}{\textbf{FT}}{\hskip 0.35in} & \multicolumn{3}{c}{\textbf{PF (5m)}}{\hskip 0.35in} & \multicolumn{3}{c}{\textbf{AP (5m)}}\\
\toprule
& & R-1 & R-2 & R-L & R-1 & R-2 & R-L & R-1 & R-2 & R-L  \\
\cmidrule{3-11}
357m & 1k & 28.52 & 8.07 & 21.97 & 30.75 & 10.19 & 24.05 & 31.51 & 10.55 & 24.47 \\
357m & 5k & 32.75 & 10.80 & 25.21 & 32.89 & 11.56 & 25.73 & 32.91 & 11.45 & 25.62 \\
357m & 10k & 34.58 & 12.54 & 27.09 & 33.25 & 11.76 & 25.94 & 33.66 & 12.05 & 26.34 \\
357m & 50k & 38.06 & 15.69 & 30.28 & 34.52 & 12.73 & 27.06 & 36.20 & 13.90 & 28.53 \\
357m & 100k & 39.56 & 17.13 & 31.73 & 34.90 & 12.93 & 27.37 & 37.37 & 14.96 & 29.49 \\
357m & 200k & 41.59 & 19.21 & 33.77 & 35.14 & 13.26 &  27.68 & 37.67 & 15.36 & 30.06 \\
\midrule
\textbf{Model} & \textbf{Samples} & \multicolumn{3}{c}{\textbf{FT}}{\hskip 0.35in} & \multicolumn{3}{c}{\textbf{PF (10m)}}{\hskip 0.35in} & \multicolumn{3}{c}{\textbf{AP (10m)}}\\
\toprule
& & R-1 & R-2 & R-L & R-1 & R-2 & R-L & R-1 & R-2 & R-L  \\
\cmidrule{3-11}
1.3b & 1k & 32.73 & 11.01 & 25.36 & 33.61 & 12.11 & 26.41 & 33.94 & 12.17 & 26.40 \\
1.3b & 5k & 34.90 & 12.49 & 26.96 & 36.34 & 14.34 & 28.71 & 36.65 & 14.61 & 28.98 \\
1.3b & 10k & 38.00 & 15.49 & 30.28 & 37.22 & 15.10 & 29.57 & 37.56 & 15.39 & 29.83 \\
1.3b & 50k & 40.92 & 18.19 & 32.89 & 38.60 & 16.38 & 30.81 & 40.09 & 17.47 & 32.08 \\
1.3b & 100k & 42.27 & 19.48 & 34.21 & 39.74 & 17.21 & 31.85 & 41.25 & 18.57 &  33.26 \\
1.3b & 200k & 43.95 & 21.06 & 35.75 & 40.38 & 17.75 & 32.37 & 42.35 & 19.55 & 34.22 \\
\midrule
\textbf{Model} & \textbf{Samples} & \multicolumn{3}{c}{\textbf{FT}}{\hskip 0.35in} & \multicolumn{3}{c}{\textbf{PF (33m)}}{\hskip 0.35in} & \multicolumn{3}{c}{\textbf{AP (33m)}}\\
\toprule
& & R-1 & R-2 & R-L & R-1 & R-2 & R-L & R-1 & R-2 & R-L  \\
\cmidrule{3-11}
8.3b & 1k & 37.23 & 14.87 & 29.45 & 37.33 & 16.17 & 29.82 & 39.32 & 17.28 & 31.23 \\
8.3b & 5k & 39.75 & 16.80 & 31.58 & 40.76 & 18.36 & 32.70 & 40.48 & 17.73 & 32.21 \\
8.3b & 10k & 40.73 & 17.76 & 32.68 & 41.78 & 19.19 & 33.60 & 41.33 & 18.60 & 33.07 \\
8.3b & 50k & 43.39 & 20.21 & 35.02 & 43.82 & 20.99 & 35.62 & 43.48 & 20.35 & 34.99 \\
8.3b & 100k & 44.55 & 21.45 & 36.13 & 44.80 & 21.94 & 36.45 & 44.26 & 21.05 & 35.80 \\
8.3b & 200k & 46.10 & 23.14 & 37.76 & 45.51 & 22.64 & 37.14 & 45.24 & 22.07 & 36.73 \\
\bottomrule
\end{tabular}
}
\caption{Full Rouge score for different model and sample size settings.}
\label{tab:add_results_1}
\end{table*}

\begin{table*}[t]
\centering
\small{
\begin{tabular}{@{}c c c c c @{\hskip 0.35in} c c c@{}}
\textbf{Model} & \textbf{Samples} & \multicolumn{3}{c}{\textbf{PF (5m)}}{\hskip 0.35in} & \multicolumn{3}{c}{\textbf{AP (5m)}}\\
\toprule
&  & R-1 & R-2 & R-L & R-1 & R-2 & R-L  \\
\cmidrule{3-8}
1.3b & 1k & 33.78 & 12.21 & 26.47 & 34.35 & 12.66,&  26.83 \\
1.3b & 5k  & 36.38 & 14.40 & 28.79 & 36.41 & 14.41  &  28.83 \\
1.3b & 10k & 37.14 &  15.08 &  29.49 & 37.41 & 15.28 & 29.71 \\
1.3b & 50k & 38.89 & 16.56 & 31.06 & 39.82 &  17.21  &  31.86 \\
1.3b & 100k & 39.78 & 17.44 & 32.01 & 40.80 & 18.15 & 32.83 \\
1.3b & 200k  & 40.62 & 18.07 & 32.76 & 41.48 & 18.97 & 33.73\\
\midrule
\textbf{Model} & \textbf{Samples} & 
\multicolumn{3}{c}{\textbf{PF (10m)}}{\hskip 0.35in} & \multicolumn{3}{c}{\textbf{AP (10m)}}\\
\toprule
&  & R-1 & R-2 & R-L & R-1 & R-2 & R-L  \\
\cmidrule{3-8}
1.3b & 1k & 33.61 & 12.11 & 26.41 & 33.94 & 12.17 & 26.40 \\
1.3b & 5k &  36.34 & 14.34 & 28.71 & 36.65 & 14.61 & 28.98 \\
1.3b & 10k & 37.22 & 15.10 & 29.57 & 37.56 & 15.39 & 29.83 \\
1.3b & 50k & 38.60 & 16.38 & 30.81 & 40.09 & 17.47 & 32.08 \\
1.3b & 100k & 39.74 & 17.21 & 31.85 & 41.25 & 18.57 &  33.26 \\
1.3b & 200k & 40.38 & 17.75 & 32.37 & 42.35 & 19.55 & 34.22 \\
\midrule
\textbf{Model} & \textbf{Samples} & 
\multicolumn{3}{c}{\textbf{PF (5m)}}{\hskip 0.35in} & \multicolumn{3}{c}{\textbf{AP (5m)}}\\
\toprule
&  & R-1 & R-2 & R-L & R-1 & R-2 & R-L  \\
\cmidrule{3-8}
8.3b & 5k  & 40.44 & 17.93 & 32.28 &  41.48 &  19.04 & 33.35 \\
8.3b & 10k  & 41.79 &  19.09 &  33.45 & 42.33 & 19.72 & 34.22 \\
8.3b & 50k  & 43.62 &  20.91 & 35.41  & 44.39 & 21.54 & 36.17\\
8.3b & 100k & 44.74 & 21.79 & 36.39 & 45.14, &  22.28 &  36.80 \\
8.3b & 200k & 45.60 &  22.68 & 37.21 & 46.04 & 23.16 & 37.71\\
\midrule
\textbf{Model} & \textbf{Samples} &
\multicolumn{3}{c}{\textbf{PF (33m)}}{\hskip 0.35in} & \multicolumn{3}{c}{\textbf{AP (33m)}}\\
\toprule
&  & R-1 & R-2 & R-L & R-1 & R-2 & R-L  \\
\cmidrule{3-8}
8.3b & 5k  & 40.76 & 18.36 & 32.70 & 40.48 & 17.73 & 32.21 \\
8.3b & 10k  & 41.78 & 19.19 & 33.60 & 41.33 & 18.60 & 33.07 \\
8.3b & 50k  & 43.82 & 20.99 & 35.62 & 43.48 & 20.35 & 34.99 \\
8.3b & 100k & 44.80 & 21.94 & 36.45 & 44.26 & 21.05 & 35.80 \\
8.3b & 200k & 45.51 & 22.64 & 37.14 & 45.24 & 22.07 & 36.73 \\
\bottomrule
\end{tabular}
}
\caption{Adding extra parameters are not always helpful. It happens across different training sample sizes.}
\label{tab:add_results_2}
\end{table*}

\end{document}